%% file: eacl2023.tex
\pdfoutput=1

\documentclass[11pt]{article}

\usepackage[]{EACL2023}

\usepackage{times}
\usepackage{latexsym}

\usepackage[T1]{fontenc}

\usepackage[utf8]{inputenc}

\usepackage{microtype}

\usepackage{dsfont}
\usepackage{bm}
\usepackage{bbm}
\usepackage{graphicx}
\usepackage{color}
\usepackage{multicol}
\usepackage{multirow}
\usepackage{wrapfig,lipsum,booktabs}
\usepackage[ruled,vlined,linesnumbered]{algorithm2e}
\usepackage{pgfplots}
\usepackage{filecontents}
\usepackage{xcolor}
\usepackage{tikz}
\usepackage{xspace}
\usepackage{makecell}
\usepackage{soul}
\usepackage{amsmath,amsfonts,amssymb}
\usepackage{subcaption}
\usetikzlibrary{calc}
\usepgfplotslibrary{groupplots}
\usetikzlibrary{angles,quotes} 
\usetikzlibrary{shapes,arrows}
\usetikzlibrary{backgrounds}
\usepackage{tikz-3dplot}
\usepackage{hyperref}
\usepackage{cleveref}
\usepackage{paralist}
\usepackage{cancel}
\usepackage{xspace}
\usepackage{todonotes}
\usepackage{tabu}
\usepackage{rotating}
\usepackage{etoolbox}
\usepackage{adjustbox}
\usepackage{enumerate}
\usepackage{enumitem}
\setitemize{noitemsep,topsep=0pt,parsep=0pt,partopsep=0pt}
\setenumerate{noitemsep,topsep=0pt,parsep=0pt,partopsep=0pt}
\usepackage{pifont}
\usepackage{cancel}

\newcommand{\cmark}{\ding{51}\xspace}%
\newcommand{\xmark}{\ding{55}\xspace}%

\newcommand{\vx}{\pmb{x}}
\newcommand{\vy}{\pmb{y}}
\newcommand{\vX}{\pmb{X}}
\newcommand{\vY}{\pmb{Y}}
\newcommand{\vC}{\pmb{C}}
\newcommand{\vg}{\pmb{g}}
\newcommand{\vB}{\pmb{B}}
\newcommand{\vH}{\pmb{H}}

\newcommand{\vQ}{\pmb{Q}}
\newcommand{\vK}{\pmb{K}}
\newcommand{\vV}{\pmb{V}}

\newcommand{\vs}{\pmb{s}}

\newcommand{\vpsi}{\pmb{\psi}}

\newcommand{\model}[1]{\textsc{#1}\xspace}
\newcommand{\ours}{\model{DocFlat}}
\newcommand{\ourscon}{\model{DocFlat\textsubscript{C}}}
\newcommand{\oursdis}{\model{DocFlat\textsubscript{D}}}
\newcommand{\oursid}{\model{DocFlat\textsubscript{I}}}
\newcommand{\docmodel}{\model{Doc2Doc}}
\newcommand{\sentmodel}{\model{Sent2Sent}}

\newcommand{\flattrans}{\model{FlatTrans}}
\newcommand{\gtrans}{\model{GTrans}}
\newcommand{\mbe}{\model{MBE}}
\newcommand{\abdmodel}{\model{ABD}}
\newcommand{\myattn}{\model{Flat-Batch Attention}}
\newcommand{\gate}{\model{Neural Context Gate}}
\newcommand{\dataattn}{\model{Attention Between Datapoints}}
\newcommand{\dataset}[1]{\texttt{#1}\xspace}

\newcommand*{\affmark}[1][*]{\textsuperscript{#1}}

\title{Document Flattening: Beyond Concatenating Context for Document-Level Neural Machine Translation}

\author{
  Minghao Wu\affmark[$\heartsuit$] \qquad George Foster\affmark[$\spadesuit$] \qquad Lizhen Qu\affmark[$\heartsuit$] \qquad Gholamreza Haffari\affmark[$\heartsuit$] \\
  \affmark[$\heartsuit$]Monash University \qquad \affmark[$\spadesuit$]Google Research \\
  \texttt{\{firstname.lastname\}@monash.edu} \qquad\texttt{fosterg@google.com}
}

\begin{document}

\renewcommand{\tableautorefname}{Table}
\renewcommand{\sectionautorefname}{Section}
\renewcommand{\subsectionautorefname}{Section}
\renewcommand{\subsubsectionautorefname}{Section}
\renewcommand{\figureautorefname}{Figure}
\renewcommand{\equationautorefname}{Equation}
\renewcommand{\algorithmautorefname}{Algorithm}
\newcommand{\linenoautorefname}{Line}

\maketitle
\begin{abstract}
Existing work in document-level neural machine translation commonly concatenates several consecutive sentences as a \textit{pseudo-document}, and then learns inter-sentential dependencies.
This strategy limits the model's ability to leverage information from distant context.
We overcome this limitation with a novel Document Flattening (\ours) technique that integrates \myattn (FBA) and \gate (NCG) into Transformer model to utilize information beyond the pseudo-document boundaries.
FBA allows the model to attend to all the positions in the batch and learns the relationships between positions explicitly and NCG identifies the useful information from the distant context.
We conduct comprehensive experiments and analyses on three benchmark datasets for English-German translation, and validate the effectiveness of two variants of \ours.
Empirical results show that our approach outperforms strong baselines with statistical significance on BLEU, COMET and accuracy on the contrastive test set.
The analyses highlight that \ours is highly effective in capturing the long-range information.
\end{abstract}

\input{1_introduction.tex}
\input{2_preliminaries.tex}

\input{3_method.tex}

\input{4_setup.tex}

\input{5_main_results.tex}

\input{6_analysis.tex}

\input{7_related_work.tex}
\input{8_conclusion.tex}

\input{9_limitation.tex}

\bibliography{anthology,custom}
\bibliographystyle{acl_natbib}

\clearpage
\appendix

\input{appendix.tex}

\end{document}

%% file: 1_introduction.tex
\section{Introduction}

Remarkable progress has been made in neural machine translation (NMT) \cite{DBLP:conf/nips/SutskeverVL14, DBLP:conf/nips/VaswaniSPUJGKP17, chen-etal-2018-best}, yet human translation still clearly outperforms NMT at the document level \cite{laubli-etal-2018-machine, freitag-etal-2021-experts}, because current sentence-level NMT systems ignore the inter-sentential relationships.
To narrow this gap, numerous document-level NMT (DocNMT) approaches have been proposed in recent years to improve the context awareness by incorporating the contextual information during the translation \cite{tiedemann-scherrer-2017-neural, maruf-haffari-2018-document, wong-etal-2020-contextual}.

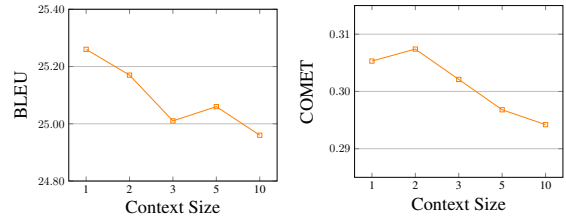
\begin{figure}[t]  
    \centering 
    \begin{subfigure}[c]{0.45\linewidth}
        \centering
        \begin{tikzpicture}[scale=0.4]
            \begin{axis}[
                xlabel={\LARGE Context Size},
                ylabel={\LARGE BLEU},
                y label style={at={(-0.05,0.5)}},
                xtick={1,2,3,4,5},
                xticklabels={1,2,3,5,10},
                ymajorgrids = true,
                grid style={line width=.1pt, draw=gray!50},
                y tick label style={
                    /pgf/number format/.cd,
                    fixed,
                    fixed zerofill,
                    precision=2
                },
                ymin=24.8, ymax=25.4,
                legend style={
                    at={(1,1)},
                    anchor=south east,
                    column sep=0ex,
                    font=\scriptsize,
                    legend columns=-1,
                    legend cell align=left,
                }
            ]
            \addplot[color=orange,mark=square] coordinates {
                (1,25.26)
                (2,25.17)
                (3,25.01)
                (4,25.06)
                (5,24.96)
            };
            \end{axis}  
        \end{tikzpicture}
    \end{subfigure}
    \hspace{1mm}
    \begin{subfigure}[c]{0.45\linewidth}
        \centering
        \begin{tikzpicture}[scale=0.4] 
            \begin{axis}[
                xlabel={\LARGE Context Size},
                ylabel={\LARGE COMET},
                y label style={at={(-0.05,0.5)}},
                xtick={1,2,3,4,5},
                xticklabels={1,2,3,5,10},
                ymin=0.285, ymax=0.315,
                ymajorgrids = true,
                grid style={line width=.1pt, draw=gray!50},
                y tick label style={
                    /pgf/number format/.cd,
                    fixed,
                    fixed zerofill,
                    precision=2
                },
                legend style={
                    at={(1,1)},
                    anchor=south east,
                    column sep=0ex,
                    font=\scriptsize,
                    legend columns=-1,
                    legend cell align=left,
                }
            ]
            \addplot[color=orange,mark=square] coordinates {
                (1,0.3053)
                (2,0.3074)
                (3,0.3021)
                (4,0.2968)
                (5,0.2942)
            };
            \end{axis} 
        \end{tikzpicture}
    \end{subfigure}
    \caption{
        The change of BLEU (left) and COMET (right) given by \docmodel on \dataset{TED} with regard to the context size of \textit{pseudo-document} (in sentences) based on the experimental setup described in \autoref{sec:setup}.
    }
    \label{fig:doc_ctx_size}
\end{figure}

Existing DocNMT systems commonly concatenate several consecutive sentences to form a \textit{pseudo-document}, instead of processing the \textit{entire document} \cite{zhang-etal-2018-improving,voita-etal-2019-good,junczys-dowmunt-2019-microsoft,fernandes-etal-2021-measuring}.
One typical pseudo-document contains the \textit{current sentence} to be translated and the \textit{surrronding context}.
Intuitively, larger context should result in better performance.
In our preliminary study, the model performance does not always grow as the context size increases as shown in \autoref{fig:doc_ctx_size}.
\citet{liu-etal-2020-multilingual-denoising} and \citet{bao-etal-2021-g} also observe that Transformer's performance declines with longer inputs.
We refer to this phenomenon as the \textit{quality saturation problem} \cite{glaser1967discovery}.
Therefore, such formation of pseudo-document limits the DocNMT systems to leverage the information from a relatively small context.
Consequently, once the entire original document is segmented into several pseudo-documents for reducing the sequence length, the information out of the pseudo-document's scope is no longer accessible to the current sentence.
Therefore, a natural research question to ask is that, \textit{is there a more effective way to model the parallel documents in DocNMT?}

In this work, we seek DocNMT approaches that could better expand the context scope and improve the corresponding translation performance. 
Instead of directly training DocNMT system on the entire document, we propose to store the document as multiple pseudo-documents in a single batch and optimize the DocNMT models by leveraging the inter-pseudo-document relationships at the batch level.
Inspired by \citet{DBLP:conf/nips/KossenBLGRG21}, we propose a Document Flattening (\ours) technique that integrates \myattn (FBA) and \gate (NCG) into the Transformer model \cite{DBLP:conf/nips/VaswaniSPUJGKP17}.
FBA flattens all the current sentences in the batch with the original order into a sequence along the temporal dimension.
It then applies the attention mechanism to the flattened sequence.
The goal of this design is to preserve the linguistic structure of documents and expand the scope of context by explicitly learning the pseudo-document relationships.
As there is both supportive and noisy information in the longer context, we introduce NCG, a simple feed-forward network, to identify the usefulness of contextual information and filter out the noise.
With the combination of FBA and NCG, \ours effectively captures the information in the distant context.
To the best of our knowledge, \citet{morishita-etal-2021-context} propose mini-batch embedding (\mbe), which is the only close work to ours.
They compute the average representation for all the source tokens in the batch and prepend it to the source and target pseudo-documents.
The compressed representation ignores the linguistic structure of documents, providing limited contextual information.

Our contributions are summarized as follows.
Firstly, we propose a novel approach \ours that allows the model to attend the content beyond the pseudo-document boundaries using FBA and NCG.
Secondly, we demonstrate that \ours outperforms strong baselines with statistical significance, in terms of BLEU, COMET and accuracy on the contrastive test set, on three DocNMT benchmark datasets, including \dataset{TED}, \dataset{News Commentary} and \dataset{Europarl}.
Thirdly, we conduct comprehensive analyses to understand the effectiveness of \ours. 
The analyses highlight that \ours is highly effective in capturing the distant context.

%% file: 2_preliminaries.tex
\section{Preliminaries}
\label{sec:prelim}

\paragraph{Sentence-level NMT (SentNMT)} 
The sentence-level NMT model neglects the inter-sentential dependencies between the current sentence and its context.
Its probability of translation is defined as:
\begin{align}
    P(\vy_{i}|\vx_{i}) = \prod^{d}_{t=1}P(y_{i,t}|\vy_{i,<t}, \vx_{i}),
\end{align}
where $\vx_{i}$ and $\vy_{i}$ are the $i$-th source and target training sentence, $y_{i,t}$ denotes the $t$-th token in $\vy_{i}$ and $d$ is the sentence length of $\vy_{i}$.

\paragraph{Document-level NMT (DocNMT)} 
Given a document pair $\{(\vx_i, \vy_i)\}^{M}_{i=1}$ where we denote the aligned sentence pair as $\vx_i$ and $\vy_i$ and $M$ is the length of document in sentences, 
the $i$-th pseudo-document pair $\vX_i$ and $\vY_i$ can be defined as:
\begin{align}
    \begin{split}
        \vX_{i}=\textrm{Concat}([\vx_{i-c^{-}}, \ldots, \vx_{i}, \ldots, \vx_{i+c^{+}}]), \\
        \vY_{i}=\textrm{Concat}([\vy_{i-c^{-}}, \ldots, \vy_{i}, \ldots, \vy_{i+c^{+}}]),
    \end{split}
\end{align}
where $c^{-}$ is the context size before the current sentence and $c^{+}$ is the context size after the current sentence.
The translation probability of target \emph{current} sentence $\vy_{i}$ in the target pseudo-document $\vY_{i}$ given the source pseudo-document $\vX_{i}$ in DocNMT can be written as:
\begin{align}
    \label{eq:docnmt}
    P(\vy_{i}|\vx_{i}, \vC_{-i}) = \prod^{d}_{t=1}P(y_{i,t}|\vy_{i,<t}, \vx_{i}, \vC_{-i}),
\end{align}
where $\vC_{-i}$ is the collection of all the sentences in the pseudo-document pair except $(\vx_{i}, \vy_{i})$, and $\vx_{i}$ is the source \emph{current} sentence.
We do not consider the context after the current sentence in this work, so $c^{+}$ is 0.

%% file: 3_method.tex
\section{Document Flattening}
\label{sec:method}
In this section, we firstly describe the overview of \ours (\autoref{sec:overview}).
We then introduce \ours's core components, \myattn (FBA; \autoref{sec:myattn}) and \gate (NCG; \autoref{sec:gate}).
Finally, we discuss the practical considerations (\autoref{sec:shuffle} and \autoref{sec:infer}) of \ours along with a concrete example.

\subsection{Overview of \ours}
\label{sec:overview}

\begin{figure}[t]
    \scalebox{0.53}{
        \centering
        \tikzstyle{embednode} = [
            rectangle, 
            rounded corners, 
            minimum width=3cm, 
            minimum height=1cm, 
            text width=3cm,
            text centered, 
            draw=black, 
            fill=green!20
        ]
        \tikzstyle{attnnode} = [
            rectangle, 
            rounded corners, 
            minimum width=3cm, 
            minimum height=1cm, 
            text width=3cm,
            text centered, 
            draw=black, 
            fill=blue!20
        ]
        \tikzstyle{ffnnode} = [
            rectangle, 
            rounded corners, 
            minimum width=3cm, 
            minimum height=1cm, 
            text width=3cm,
            text centered, 
            draw=black, 
            fill=red!20
        ]
        \tikzstyle{softmaxnode} = [
            rectangle, 
            rounded corners, 
            minimum width=3cm, 
            minimum height=1cm, 
            text width=3cm,
            text centered, 
            draw=black, 
            fill=yellow!20
        ]
    
        \tikzstyle{norm} = [
            rectangle, 
            rounded corners, 
            minimum width=2cm, 
            minimum height=0.5cm, 
            text width=2cm,
            text centered, 
            draw=black, 
            fill=red!40
        ]
    
        \tikzstyle{fbanode} = [
            rectangle, 
            rounded corners, 
            minimum width=2cm, 
            minimum height=1cm, 
            text width=2cm,
            text centered, 
            draw=black, 
            fill=blue!40
        ]
        \tikzstyle{ncgnode} = [
            rectangle, 
            rounded corners, 
            minimum width=2cm, 
            minimum height=1cm, 
            text width=2cm,
            text centered, 
            draw=black, 
            fill=orange!40
        ]
    
        \tikzstyle{decoder} = [
            rectangle, 
            minimum width=4cm, 
            minimum height=4.5cm, 
            text width=3cm,
            fill=blue!10,
        ]
        \tikzstyle{encoder} = [
            rectangle, 
            minimum width=4cm, 
            minimum height=3cm, 
            text width=3cm,
            fill=blue!10,
        ]
        \tikzstyle{textnode} = [
            rectangle, 
            minimum width=1.5cm, 
            minimum height=0.5cm, 
            text width=1.5cm,
            text centered, 
        ]
        \tikzstyle{circ} = [
            circle, 
            minimum width=0.5cm, 
            minimum height=0.5cm, 
            text centered, 
            draw=black,
        ]

        \begin{tikzpicture}[node distance=2cm]
        
            \node (softmax) [softmaxnode] {Softmax};
            \node (decffn) [ffnnode, below=0.5cm of softmax] {Feed-Forward};
            \node (crossattn) [attnnode, below=0.5cm of decffn] {Cross Attention};
            \node (decselfattn) [attnnode, below=0.5cm of crossattn] {Masked Self-Attention};
            
            \node (encselfattn) [attnnode, left=3cm of decselfattn] {Self-Attention};
            \node (encffn) [ffnnode, left=3cm of crossattn] {Feed-Forward};
            \node (encnorm) [norm, below=0.5cm of encselfattn] {\small LayerNorm};
            \node (decnorm) [norm, below=0.5cm of decselfattn] {\small LayerNorm};

            
            \node (encplus) [circ,  below=0.5cm of encnorm] {};
            \draw (encplus.north) -- (encplus.south) (encplus.west) -- (encplus.east);
            \node (enctimes) [circ, left=0.5cm of encplus] {};
            \draw (enctimes.north west) -- (enctimes.south east) (enctimes.south west) -- (enctimes.north east);
            \node (srctimes) [circ, below=0.5cm of encplus] {};
            \draw (srctimes.north west) -- (srctimes.south east) (srctimes.south west) -- (srctimes.north east);

            \node (encncg) [ncgnode, left=0.5cm of enctimes] {NCG};
            \node (encfba) [fbanode, below=0.5cm of encncg] {FBA};

            \node (srcembed) [embednode, below=2.5cm of encplus] {Source Embedding};


            \node (decplus) [circ,  below=0.5cm of decnorm] {};
            \draw (decplus.north) -- (decplus.south) (decplus.west) -- (decplus.east);
            \node (dectimes) [circ, left=0.5cm of decplus] {};
            \draw (dectimes.north west) -- (dectimes.south east) (dectimes.south west) -- (dectimes.north east);
            \node (tgttimes) [circ, below=0.5cm of decplus] {};
            \draw (tgttimes.north west) -- (tgttimes.south east) (tgttimes.south west) -- (tgttimes.north east);
            
            \node (decncg) [ncgnode, left=0.5cm of dectimes] {NCG};
            \node (decfba) [fbanode, below=0.5cm of decncg] {FBA};

            \node (tgtembed) [embednode, below=2.5cm of decplus] {Target Embedding};

            \begin{scope}[on background layer]
                \node (decoder) [decoder, below=-1.25cm of decffn] {};
                \node (encoder) [encoder, below=-1.25cm of encffn] {};
            \end{scope}
            \node (enclayer) [textnode,  right=0cm of encoder] {$\times L$};
            \node (declayer) [textnode,  right=0cm of decoder] {$\times L$};

            \draw [-latex] (encselfattn) -- (encffn);
            \draw [-latex] (encnorm) -- (encselfattn);
            \draw [-latex] (encplus) -- (encnorm);
            \draw [-latex] (encfba) -- (encncg);
            \draw [-latex] (encncg) -- (enctimes);
            \draw [-latex] (enctimes) -- (encplus);
            \draw [-latex] (srcembed) -- (srctimes);
            \draw [-latex] (srctimes) -- (encplus);
            \draw [-latex] (srcembed) |- ++(0,8mm) -| (encfba.south);
            \draw [-latex] (encfba.east) -| (enctimes.south);
            \draw [-latex] (encncg.east) -| ++(2mm,0) |- (srctimes.west);

            \draw [-latex] (decselfattn) -- (crossattn);
            \draw [-latex] (crossattn) -- (decffn);
            \draw [-latex] (decffn) -- (softmax);
            \draw [-latex] (decnorm) -- (decselfattn);
            \draw [-latex] (decplus) -- (decnorm);
            \draw [-latex] (decfba) -- (decncg);
            \draw [-latex] (decncg) -- (dectimes);
            \draw [-latex] (dectimes) -- (decplus);
            \draw [-latex] (tgtembed) -- (tgttimes);
            \draw [-latex] (tgttimes) -- (decplus);
            \draw [-latex] (tgtembed) |- ++(0,8mm) -| (decfba.south);
            \draw [-latex] (decfba.east) -| (dectimes.south);
            \draw [-latex] (decncg.east) -| ++(2mm,0) |- (tgttimes.west);

            \draw [-latex] (encffn) -- (crossattn);

        \end{tikzpicture}

    }
    \caption{
        The model architecture of \ours.
        $\otimes$ denotes the element-wise multiplication.\
        $\oplus$ denotes the element-wise addition.
        More details are in \autoref{sec:method}.
    }
    \label{fig:model}
\end{figure}
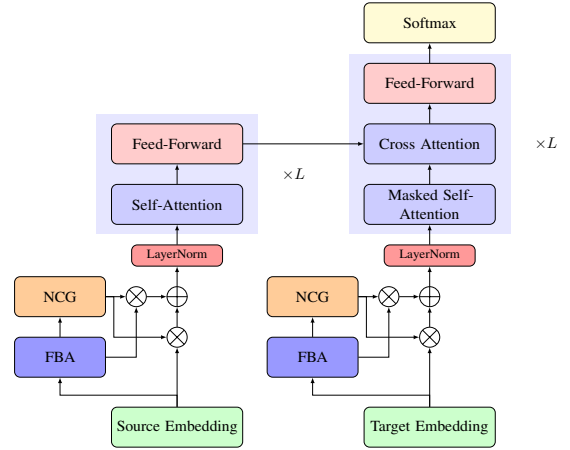

We present the overall architecture of \ours in \autoref{fig:model}.
Given a sequence-to-sequence Transformer with $L$ encoder layers and $L$ decoder layers, we apply the FBA and NCG to the input word embeddings with the residual connection and Layer Normalization, instead of directly feeding the embeddings into either encoder or decoder.
\ours's translation probability of the $i$-th target current sentence $\vy_{i}$ of the original document in the $i$-th target pseudo-document $\vY_{i}$ given the $i$-th source pseudo-document $\vX_{i}$ in the batch $\vB=\{(\vX_{j}, \vY_{j})\}_{j=1}^{n}$, where $n$ is the batch size, is defined as:
\begin{align}
    \label{eq:loss}
    \begin{split}
        P(\vy_{i}|\vx_{i}, &\vC_{-i}, \vB_{-i}) = \\ &\prod^{d}_{t=1}P(y_{i,t}|\vy_{i,<t}, \vx_{i}, \vC_{-i}, \vB_{-i}),
    \end{split}
\end{align}
where $\vC_{-i}$ is defined as in \autoref{eq:docnmt} and $\vB_{-i}$ is the collection of all the current sentences in the batch except $(\vx_{i}, \vy_{i})$.
We categorize the context for the current sentence into two groups, the global context (GC) from other pseudo-documents $\vB_{-i}$ and the local context (LC) from its own pseudo-document $\vC_{-i}$ (See \autoref{fig:exmaple}).

\subsection{\myattn}
\label{sec:myattn}

\paragraph{Multi-Head Self-Attention (MHSA)}
Scaled dot-product attention is the core mechanism of Transformer model with the inputs of query $\vQ$, key $\vK$ and $\vV$ \cite{DBLP:conf/nips/VaswaniSPUJGKP17}.
The attention mechanism computes the attention weights by comparing queries $\vQ$ with keys $\vK$ and then updates the representations of queries by computing the weighted sum of values $\vV$ with the attention weights, which is described as follows:
\begin{align}
    \label{eq:attn}
    \textrm{Attn}(\vQ, \vK, \vV) = \textrm{softmax}(\frac{\vQ\vK^{\top}}{\sqrt{e}}) \vV,
\end{align}
where $e$ is the hidden state dimension.
Multi-head self-attention (MHSA) then allows the model to jointly attend to information from different hidden subspaces by concatenating a sequence of independent attention heads as follows:
\begin{align}
    \label{eq:mhsa}
    \begin{split}
        \textrm{MHSA}(\vQ, &\vK, \vV) =\\
        &\textrm{Concat}(\textrm{head}_{1}, \ldots, \textrm{head}_{k}),
    \end{split}
\end{align}
where $\textrm{head}_{j}$ is the scaled dot-product attention in \autoref{eq:attn} with independent parameters and $j \in \{1, \ldots, k\}$ for each head $j$.

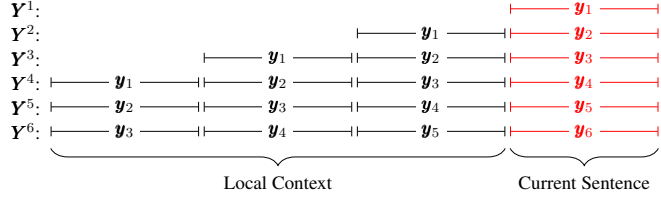
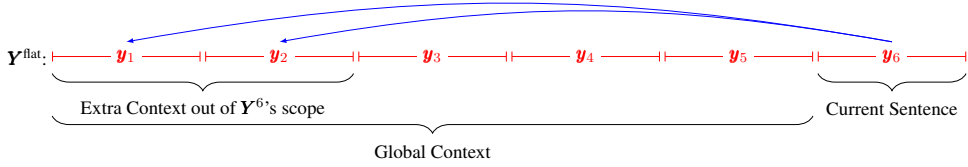
\begin{figure*}[t]
    
    \begin{subfigure}{\textwidth}
        \centering
        \scalebox{0.65}{
            \begin{tikzpicture}

                \node[] at (-0.5, 0) () {$\vY^{1}$:} ;
                \draw[|-|, red] (9.3,0) --  node[fill=white] {$\vy_{1}$} (12.3,0);
    
                \node[] at (-0.5, -0.5) () {$\vY^{2}$:} ;
                \draw[|-|] (6.2,-0.5) --  node[fill=white] {$\vy_{1}$} (9.2,-0.5);
                \draw[|-|, red] (9.3,-0.5) --  node[fill=white] {$\vy_{2}$} (12.3,-0.5);
    
                \node[] at (-0.5, -1) () {$\vY^{3}$:} ;
                \draw[|-|] (3.1,-1) --  node[fill=white] {$\vy_{1}$} (6.1,-1);
                \draw[|-|] (6.2,-1) --  node[fill=white] {$\vy_{2}$} (9.2,-1);
                \draw[|-|, red] (9.3,-1) --  node[fill=white] {$\vy_{3}$} (12.3,-1);
    
                \node[] at (-0.5, -1.5) () {$\vY^{4}$:} ;
                \draw[|-|] (0,-1.5) --  node[fill=white] {$\vy_{1}$} (3,-1.5);
                \draw[|-|] (3.1,-1.5) --  node[fill=white] {$\vy_{2}$} (6.1,-1.5);
                \draw[|-|] (6.2,-1.5) --  node[fill=white] {$\vy_{3}$} (9.2,-1.5);
                \draw[|-|, red] (9.3,-1.5) --  node[fill=white] {$\vy_{4}$} (12.3,-1.5);
    
                \node[] at (-0.5, -2) () {$\vY^{5}$:} ;
                \draw[|-|] (0,-2) --  node[fill=white] {$\vy_{2}$} (3,-2);
                \draw[|-|] (3.1,-2) --  node[fill=white] {$\vy_{3}$} (6.1,-2);
                \draw[|-|] (6.2,-2) --  node[fill=white] {$\vy_{4}$} (9.2,-2);
                \draw[|-|, red] (9.3,-2) --  node[fill=white] {$\vy_{5}$} (12.3,-2);
    
                \node[] at (-0.5, -2.5) () {$\vY^{6}$:} ;
                \draw[|-|] (0,-2.5) --  node[fill=white] {$\vy_{3}$} (3,-2.5);
                \draw[|-|] (3.1,-2.5) --  node[fill=white] {$\vy_{4}$} (6.1,-2.5);
                \draw[|-|] (6.2,-2.5) --  node[fill=white] {$\vy_{5}$} (9.2,-2.5);
                \draw[|-|, red] (9.3,-2.5) --  node[fill=white] {$\vy_{6}$} (12.3,-2.5);

                \draw [decorate,decoration={brace,amplitude=10pt,mirror,raise=2ex}]
                    (0,-2.5) -- (9.2,-2.5) node[midway,yshift=-30pt]{Local Context};
                \draw [decorate,decoration={brace,amplitude=10pt,mirror,raise=2ex}]
                    (9.3,-2.5) -- (12.3,-2.5) node[midway,yshift=-30pt]{Current Sentence};

            \end{tikzpicture}
        }
        \caption{
            An example batch of pseudo-documents $\vB_{\textrm{tgt}}=\{ \vY^{1}, \vY^{2}, \vY^{3}, \vY^{4}, \vY^{5}, \vY^{6} \}$ at the target side.
            Each $\vY^{j}$ contains four consecutive sentences and $\vy_{i}$ indicates the $i$-th sentence of the same original document.
            \textcolor{red}{The segments in red} indicate the current sentence of each pseudo-document.
        }
        \label{fig:pseudo_docs}
    \end{subfigure}
    \vfill
    \begin{subfigure}{\textwidth}
        \centering
        \scalebox{0.65}{
            \begin{tikzpicture}
                \node[] at (-0.5, 0) () {$\vY^{\textrm{flat}}$:} ;
                \draw[|-|,red] (0,0) --  node[fill=white] {$\vy_{1}$} (3,0);
                \draw[|-|,red] (3.1,0) --  node[fill=white] {$\vy_{2}$} (6.1,0);
                \draw[|-|,red] (6.2,0) --  node[fill=white] {$\vy_{3}$} (9.2,0);
                \draw[|-|,red] (9.3,0) --  node[fill=white] {$\vy_{4}$} (12.3,0);
                \draw[|-|,red] (12.4,0) --  node[fill=white] {$\vy_{5}$} (15.4,0);
                \draw[|-|,red] (15.5,0) --  node[fill=white] {$\vy_{6}$} (18.5,0);
    
                \draw [-latex,blue] (17,0.3) to [out=170,in=10] (1.5,0.3);
                \draw [-latex,blue] (17,0.3) to [out=170,in=10] (4.6,0.3);

                \draw [decorate,decoration={brace,amplitude=10pt,mirror,raise=2ex}]
                    (0,0) -- (6.1,0) node[midway,yshift=-30pt]{Extra Context out of $\vY^{6}$'s scope};
                \draw [decorate,decoration={brace,amplitude=10pt,mirror,raise=7ex}]
                    (0,0) -- (15.4,0) node[midway,yshift=-55pt]{Global Context};
                \draw [decorate,decoration={brace,amplitude=10pt,mirror,raise=2ex}]
                    (15.5,0) -- (18.5,0) node[midway,yshift=-30pt]{Current Sentence};
    
            \end{tikzpicture}
        }
        \caption{
            An example of the flattened sequence $\vY^{\textrm{flat}}$ transformed from $\vB_{\textrm{tgt}}$ in \autoref{fig:pseudo_docs} with \myattn.
            For the current sentence $\vy_{6}$, \textcolor{blue}{The blue arrows} indicate the extra inter-sentential attention for $\vy_{6}$ that our approach can model.
            $\vy_{1}$ and $\vy_{2}$ are the extra context introduced by our approach. 
        }
        \label{fig:flattenseq}
    \end{subfigure}
    \caption{
        An example batch of pseudo-documents at the target side and its flattened sequence.
        Another example at the source side can be found at \autoref{sec:src_example}.
    }
    \label{fig:exmaple}
\end{figure*}

\paragraph{\myattn (FBA)}
To leverage the contextual information beyond the pseudo-document boundaries, we propose \myattn (FBA).
It explicitly transforms the stacked instances in the batch to a single flattened sequence of tokens as shown in \autoref{fig:exmaple}. 
Given a batch of hidden representations $\vH \in \mathbb{R}^{n \times d \times e}$ consisting of $n$ instances padded to the length of $d$ with the hidden dimension of $e$, FBA operates as follows:
\begin{align}
    \label{eq:myattn}
    \begin{split}
        \hat{\vH}_{\textrm{flat}} &= \textrm{Flatten}(\vH) \in \mathbb{R}^{(n \times d) \times e}, \\
        \hat{\vH}_{\textrm{mhsa}} &= \textrm{MHSA}(\hat{\vH}_{\textrm{flat}}, \hat{\vH}_{\textrm{flat}}, \hat{\vH}_{\textrm{flat}}), \\
        \hat{\vH}_{\textrm{rsh}} &= \textrm{Reshape}(\hat{\vH}_{\textrm{mhsa}}) \in \mathbb{R}^{n \times d \times e}, \\
        \hat{\vH} &= \textrm{LN}((1-\vg)\otimes \vH + \vg\otimes \hat{\vH}_{\textrm{rsh}}).
    \end{split}
\end{align}
As shown in \autoref{eq:myattn}, we first flatten $\vH \in \mathbb{R}^{n \times d \times e}$ to $\hat{\vH}_{\textrm{flat}} \in \mathbb{R}^{(n \times d) \times e}$, where $(n \times d)$ indicates the flattened sequence length.
The $\hat{\vH}_{\textrm{flat}}$ is then fed into a MHSA layer and reshaped back to $\hat{\vH}_{\textrm{rsh}} \in \mathbb{R}^{n \times d \times e}$.
We then add a residual connection with $\vg$ given by NCG $\vpsi$ followed by a sigmoid function $\sigma$ and apply the Layer Normalization (LN; \citealp{DBLP:journals/corr/BaKH16}) following the reshape operation.
$\otimes$ denotes the element-wise multiplication.
We discuss the details of NCG in \autoref{sec:gate}.
Note that FBA at the decoder side is associated with a causal mask to preserve the auto-regressive property. 
By attending to all the other current sentences in the batch, FBA effectively allows the current sentences to access a much larger context than the self-attention on the pseudo-documents. 
In addition, this does not increase the input length of each instance, preventing the quality saturation problem as shown in \autoref{fig:doc_ctx_size}.

\paragraph{Complexity}
Given a Transformer model with $L$ encoder layers and $L$ decoder layers, suppose the average sentence length is $n$, the pseudo-document contains $c$ consecutive sentences, and the batch size is $b$.
The complexity of self-attention layer in the concatenation-based \docmodel is $\mathcal{O}(L(cn)^{2})$.
The extra complexity introduced by FBA is $\mathcal{O}((bn)^{2})$.
$Lc^{2}$ and $b^{2}$ have the same order of magnitude.
The batch size $b$ is set to be constant in practice and the self-attention operation in FBA is highly parallelizable, so integrating FBA into Transformer does not significantly increase the computational cost.
Empirically, \ours is only 3\% slower in training and 15\% slower in inference, compared with \docmodel (See \autoref{sec:main_results}).

\subsection{\gate}
\label{sec:gate}

The distant context can contain both supportive and noisy information.
Supportive information can assist the translation of the current sentence, while the noise may damage the model predictions.
To address this issue, we introduce a novel \gate (NCG) to automatically identify the context usefulness and control the information flow from the distant context.

In this work, NCG $\vpsi$ is a single-layer element-wise feed-forward neural network followed by a sigmoid function $\sigma$.
Given a batch of hidden representations $\vH$, the operations are defined as follows:
\begin{align}
    \label{eq:con_gate}
    \begin{split}
        \vg &= \sigma(\vpsi(\textrm{FBA}(\vH))), \\
        \vH^{o} &= (1-\vg) \otimes  \vH + \vg \otimes \textrm{\textrm{FBA}}(\vH),
    \end{split}
\end{align}
where $\vg$ is the information gate given by NCG $\vpsi$ and the sigmoid function $\sigma$, $\vH^{o}$ is output of the residual connection and $\otimes$ denotes the element-wise multiplication.
The values of $\vg$ are continuous, so we denote \ours with NCG described in \autoref{eq:con_gate} as \ourscon.

However, the continuous gate may result in the noise leakage.
In a long document, the noise at different positions may accumulate, even if they are only associated with very small gating values.
The accumulated noise can make a substantial negative impact on the model predictions.
Hence, we propose the discrete NCG as follows:
\begin{align}
    \label{eq:dis_gate}
    \begin{split}
        \vg_{\textrm{D}} &= \mathbbm{1}_{\gamma}(\sigma(\vpsi(\textrm{FBA}(\vH)))), \\
        \vH^{o} &= (1-\vg_{\textrm{D}}) \otimes  \vH + \vg_{\textrm{D}} \otimes \textrm{\textrm{FBA}}(\vH),
    \end{split}
\end{align}
where $\mathbbm{1}_{\gamma}(\cdot)$ is indicator function defined as:
\begin{align}
    \mathbbm{1}_{\gamma}(g) =
        \begin{cases}
            1 & \textrm{if $g \ge \gamma$,} \\
            0 & \textrm{otherwise,}
        \end{cases}
\end{align}
where $\gamma$ is the threshold for binarizing the gating values.
We denote \ours with the discrete NCG in \autoref{eq:dis_gate} as \oursdis and set $\gamma=0.5$ in this work.
We expect \oursdis is more robust against the noise in the context.

\subsection{Data Shuffling}
\label{sec:shuffle}
\ours aims to leverage the distant context with FBA by flattening a batch of sequences into a single sequence.
As the ordering information among sentences is critical in DocNMT, we do not shuffle the pseudo-documents during the training and inference to preserve the linguistic structure of the original document.
For each sentence, we replace the \texttt{<BOS>} symbol with its global index $i$ in the document to preserve the ordering information.
\autoref{fig:pseudo_docs} is an example at the target side to demonstrate how the pseudo-documents in the batch is organized in this work.
\autoref{fig:flattenseq} demonstrates how FBA flattens a batch of pseudo-documents.
Since the pseudo-documents are not shuffled, $\vy_{6}$ can attend to $\vy_{1}$ and $\vy_{2}$ which are not in the pseudo-document $\vY^{6}$.
We apply the causal mask to FBA at the decoder side for preserving the auto-regressive property.
Note that the pseudo-documents in a batch are mostly from the same original document.
The batches crossing the document boundaries are relatively rare and have little effect on performance in our preliminary study.

\begin{table}[t]
    \small
    \centering
    \begin{tabular}{@{}lccc@{}}
    \toprule
          & Train  & Valid & Test \\ \midrule
    \dataset{TED} & $204.4K/1.7K$ & $8.9K/93$  & $2.2K/23$ \\
    \dataset{News} & $242.4K/6.1K$ & $2.3K/81$  & $3.2K/155$ \\ 
    \dataset{Europarl} & $1.8M/117.9K$ & $3.8K/240$  & $5.5K/360$ \\ \bottomrule
    \end{tabular}
    \caption{
        The number of sentences/documents of each split of the parallel corpora.
    }
    \label{tab:datastat}
\end{table}

\begin{table*}[t]
    \small
    \centering
    \setlength{\tabcolsep}{2.5pt}
    \begin{tabular}{@{}lcccccccccc@{}}
        \toprule
                                                       & \multicolumn{3}{c}{\dataset{TED}}                 & \multicolumn{3}{c}{\dataset{News}}                & \multicolumn{3}{c}{\dataset{Europarl}}      & \multirow{2}{*}{UPS}  \\ \cmidrule(rl){2-4} \cmidrule(rl){5-7} \cmidrule(rl){8-10}
                                                       & BLEU           & COMET           & Acc.           & BLEU           & COMET           & Acc.           & BLEU           & COMET           & Acc.     &      \\ \midrule
        \textit{Reported}                              &                &                 &                &                &                 &                &                &                 &          &    \\
        DocTransformer \cite{zhang-etal-2018-improving} & 24.00          & ---              & ---             & 23.08          & ---              & ---             & 29.32          & ---              & ---       & ---    \\
        HAN \cite{miculicich-etal-2018-document}       & 24.58          & ---              & ---             & 25.03          & ---              & ---             & 28.60          & ---              & ---       & ---    \\
        Selective \cite{maruf-etal-2019-selective}     & 24.42          & ---              & ---             & 24.84          & ---              & ---             & 29.75          & ---              & ---       & ---    \\
        Hybrid \cite{DBLP:conf/ijcai/ZhengYHCB20}      & 25.10          & ---              & ---             & 24.91          & ---              & ---             & 30.40          & ---              & ---       & ---    \\ \midrule
        \textit{Re-produced (standard)}                &                &                 &                &                &                 &                &                &                 &          &      \\
        \sentmodel                                     & 24.78          & 0.2860          & 46.48          & 25.00          & 0.1993          & 47.71          & 31.24          & 0.5933          & 53.02    & 2.57     \\
        \docmodel                                      & 25.01          & 0.3021          & 66.99          & 24.95          & 0.1990          & 64.21          & 31.65          & 0.5929          & 78.18    & 0.86     \\
        \flattrans                                     & 24.71          & 0.2963          & 45.45          & 25.05          & 0.2020          & 48.54          & 31.58          & 0.5954          & 51.14    & 0.90     \\
        \gtrans                                        & 25.29          & 0.3058          & ---             & 25.59          & 0.2097          & ---             & 32.33          & 0.5904          & ---       & ---    \\ \midrule
        \textit{Re-produced (batch-level)}             &                &                 &                &                &                 &                &                &                 &          &      \\
        \mbe                                           & 24.75          & 0.3032          & 68.12          & 24.86          & 0.1969          & 62.82          & 31.63          & 0.5954          & 77.08    & 0.64     \\
        \abdmodel                                      & 24.97          & 0.3046          & 68.25          & 24.33          & 0.1772          & 62.52          & 31.98          & 0.5955          & 78.16    & 0.84     \\ \midrule
        \textit{Ours}                                  &                &                 &                &                &                 &                &                &                 &          &      \\
        \ourscon                                       & 25.31\textsuperscript{\rlap{$\dagger$}}          & \textbf{0.3173}\textsuperscript{\rlap{$\dagger$}}  & 70.92\textsuperscript{\rlap{$\dagger$}}          & \textbf{25.96}\textsuperscript{\rlap{$\dagger$}} & \textbf{0.2199}\textsuperscript{\rlap{$\dagger$}} & 65.45\textsuperscript{\rlap{$\dagger$}}          & \textbf{32.38}\textsuperscript{\rlap{$\dagger$}} & \textbf{0.6020}\textsuperscript{\rlap{$\dagger$}} & 77.65    &  0.84    \\
        \oursdis                                       & \textbf{25.41}\textsuperscript{\rlap{$\dagger$}}  & 0.3101\textsuperscript{\rlap{$\dagger$}}          & \textbf{72.04}\textsuperscript{\rlap{$\dagger$}} & 25.38          & 0.2119\textsuperscript{\rlap{$\dagger$}}          & \textbf{66.70}\textsuperscript{\rlap{$\dagger$}} & 32.16\textsuperscript{\rlap{$\dagger$}}          & 0.5990\textsuperscript{\rlap{$\dagger$}}          & \textbf{79.68}\textsuperscript{\rlap{$\dagger$}}  & 0.84 \\ \bottomrule
    \end{tabular}
    \caption{
        BLEU, COMET and accuracy on three benchmark datasets for English-German translation.
        UPS ($\uparrow$) indicates updates per second.
        The best results are highlighted in \textbf{bold}.
        --- indicate the result is not available.
        $\dagger$ indicates the statistical significance at $p=0.05$ against re-implemented \docmodel based on \citet{koehn-2004-statistical}.
    }
    \label{tab:main}
\end{table*}

\subsection{Inference}
\label{sec:infer}
We discuss the batch inference of \ours in this section.
At the encoder side, each source current sentence can attend to its own local context (LC) and all other source current sentence as the global context (GC) in the batch during the inference, as it is at the training stage.
At the decoder side, all the target current sentences are translated simultaneously, so each target current sentence can attend to its own LC and partially translated target GC.
For example, all the target current sentences in \autoref{fig:pseudo_docs} are translated simultaneously during the inference.
$\vy_{6}$ is conditioned on its own LC, $\vy_{3}$, $\vy_{4}$ and $\vy_{5}$, and partially translated target GC.
Additionally, we use the batched inference as usual and there is no overlap between batches.
For example, the first batch of sentences to be translated is $\{ \vy_{1}, \cdots, \vy_{b} \}$, and the second batch of sentences to be translated is $\{ \vy_{b+1}, \cdots, \vy_{2b} \}$, where $b$ is the inference batch size.
For decoding, we used the iterative decoding method for decoding \citep{maruf-haffari-2018-document, maruf-etal-2019-selective}. 
The initial translations of each sentence were generated by a SentNMT model, and then, we translate each sentence using the \model{DocNMT} model with the translations in the first pass as the context.

%% file: 4_setup.tex
\section{Experiments}
\label{sec:experiments}

\subsection{Setup}
\label{sec:setup}

\paragraph{Datasets}
We conduct experiments on three benchmark datasets for English-German translation, including the small-scale datasets \dataset{TED} \cite{cettolo-etal-2012-wit3} and \dataset{News Commentary} \cite{tiedemann-2012-parallel}, and the large-scale dataset \dataset{Europarl} \cite{koehn-2005-europarl}.
We tokenize the datasets with the Moses \cite{koehn-etal-2007-moses} and apply BPE \cite{sennrich-etal-2016-neural} with $32K$ merges.
Data statistics can be found in \autoref{tab:datastat}. 
We choose up to 3 previous sentences as the local context for each source and target sentence to form the pseudo-document unless otherwise specified.

\paragraph{Evaluation} 
We report the detokenized BLEU \cite{papineni-etal-2002-bleu} using SacreBLEU \cite{post-2018-call} and the neural-based COMET \cite{rei-etal-2020-comet} to measure the translation quality.\footnote{SacreBLEU Signature: \texttt{nrefs:1|case:mixed|\\eff:no|tok:13a|smooth:exp|version:2.2.0} and COMET Signature: \texttt{wmt20-comet-da}.}
We report the results with inference batch size of 16 and beam size of 5 for all the approaches, unless otherwise specified.

\paragraph{Contrastive Evaluation}
This evaluation paradigm is proposed to evaluate the contextual awareness of \model{DocNMT} models with an independent test set,
where each test example includes one correct translation and several incorrect translations.
The model is required to identify the correct translation and its overall performance is measured by micro-average \textit{Accuracy}.
In this work, we use the large-scale English-German anaphoric pronoun test set from \citet{muller-etal-2018-large}, 
containing $12K$ contrastive examples.
Given the provided context, the model of interest is required to identify the translation with the correct use of pronoun from \textit{er}, \textit{es} and \textit{sie} in German.

\paragraph{Models}
All the models in this work are based on the standard Transformer base \cite{DBLP:conf/nips/VaswaniSPUJGKP17}.
Besides the direct comparisons with prior works, we also compare \ours with several re-implemented baselines, including \sentmodel \cite{DBLP:conf/nips/VaswaniSPUJGKP17}, \docmodel \cite{tiedemann-scherrer-2017-neural}, \flattrans \cite{ma-etal-2020-simple}, \mbe \cite{morishita-etal-2021-context} and \abdmodel \cite{DBLP:conf/nips/KossenBLGRG21}.
We only apply \abdmodel at the encoder side in this work unless otherwise specified, which is its best-performing setup as shown in \autoref{sec:abd}.
We re-produce the results of \gtrans \cite{bao-etal-2021-g} with its official code and recommended hyperparameters. 
The optimization details are in \autoref{sec:hyperparam}.

%% file: 5_main_results.tex
\subsection{Main Results}
\label{sec:main_results}

We present the main results in \autoref{tab:main}.

\paragraph{Comparisons with Baselines}
Compared with all the baselines regardless whether they utilize the batch information or not, both \ourscon and \oursdis substantially outperform these strong baseline approaches, especially in terms of the context awareness (accuracy) which is the main emphasis of this work.
For the approaches that utilize the batch information, we observe that \mbe and \abdmodel only marginally improves the performance compared with \docmodel, suggesting the importance of preserving the linguistic structure in utilizing the batch-level information for DocNMT.
We also observe the larger performance gain from \ours on small \dataset{TED} and \dataset{News}, implying \ours performs better in the low-resource settings.

\paragraph{\ourscon vs. \oursdis}
As shown in \autoref{tab:main}, 
\ourscon and \oursdis demonstrate different strengths:
\ourscon mainly improves the translation quality (BLEU and COMET), while \oursdis improves the context awareness (accuracy).
In the contrastive evaluation, we have no access to the entire document, so the model predictions are always conditioned on the golden local context (LC) and irrelevant global context (GC).
\oursdis outperforms \ourscon in terms of accuracy, suggesting the discrete NCG is more robust against the noise in the context as we expected in \autoref{sec:gate}.
However, \oursdis also aggressively filters out the supportive information in the context as demonstrated on its lower results in BLEU and COMET on \dataset{News} and \dataset{Europarl}.
We believe tuning $\gamma$ in \autoref{eq:dis_gate} can fix this issue.

\paragraph{Computational Efficiency}
As described in \autoref{sec:myattn}, FBA introduces additional computational overhead.
We thus evaluate computational efficiency of \ours along with the baselines in terms of update per second (UPS) and report the results in \autoref{tab:main}.
When the context size of the pseudo-document is the same, our approach \ours is almost as fast as the standard \docmodel on the identical computational infrastructure (one Tesla A40 GPU) with significant performance gain.
Note that \gtrans \cite{bao-etal-2021-g} does not support FP16 mode, so its UPS is not reported.
During the inference, \ours is only 15\% slower than \docmodel.

\begin{table}[t]
    \small
    \centering
    \setlength{\tabcolsep}{4pt}
    \begin{tabular}{@{}lccccc@{}}
    \toprule
              & Enc.   & Dec.   & BLEU  & COMET  & Acc.  \\ \midrule
    \docmodel & $\varnothing$ & $\varnothing$ & 24.86 & 0.2821 & 66.98 \\ \midrule
    \ourscon  &        &        & 25.31 & 0.3173 & 70.92 \\
              &        & $\varnothing$ & 25.58 & 0.3114 & \textbf{71.83} \\
              & $\varnothing$ &        & \textbf{25.70} & \textbf{0.3176} & 71.06 \\ \midrule
    \oursdis  &        &        & 25.41 & 0.3101 & 72.04 \\
              &        & $\varnothing$ & 25.22 & 0.3112 & \textbf{73.60} \\
              & $\varnothing$ &        & \textbf{25.71} & \textbf{0.3113} & 70.48 \\ \bottomrule
    \end{tabular}
    \caption{
        Ablation study for FBA on \dataset{TED}.
        $\varnothing$ indicates FBA is removed. 
        The best results for \ourscon and \oursdis are highlighted in \textbf{bold} respectively.
    }
    \label{tab:usage_fba_ted}
\end{table}

\subsection{Ablation Study}
\begin{table}[t]
    \small
    \centering
    \begin{tabular}{@{}lccc@{}}
    \toprule
            & BLEU  & COMET  & Acc.  \\ \midrule
    \ourscon & 25.31 & \textbf{0.3173} & 70.92 \\
    \oursdis & \textbf{25.41} & 0.3101 & \textbf{72.04} \\
    \oursid  & 25.07 & 0.3049 & 69.45 \\ \bottomrule
    \end{tabular}
    \caption{
        Ablation study for NCG on \dataset{TED}.
        \oursid indicates \ours with identity mapping in NCG.
        The best results are highlighted in \textbf{bold}.
    }
    \label{tab:ncg}
\end{table}
\paragraph{Ablation Study for FBA}
We conduct the ablation study for FBA and present the results in \autoref{tab:usage_fba_ted}.
Compared with \docmodel, FBA at either side can effectively improve the model performance of \ours in terms of BLEU, COMET and accuracy, although both FBAs does not demonstrate orthogonal effectiveness.
We also observe that, when the FBA at the decoder side is removed, the contextual awareness (accuracy) is slightly improved.
All these results demonstrate that FBA can effectively leverage the distant context beyond the pseudo-document boundaries.

\paragraph{Ablation Study for NCG}
We present the ablation study for NCG in \autoref{tab:ncg}.
To probe the utility of NCG, we replace NCG in \autoref{eq:myattn} with identity mapping \cite{DBLP:conf/cvpr/HeZRS16} and denote this variant of \ours as \oursid.
The results from \oursid support our argument that not all the information in the context is useful.
Both \ourscon and \oursdis outperform \oursid on BLEU, COMET and accuracy, which confirms NCG can effectively filter out the noise in the context.

\begin{table}[t]
    \small
    \centering

    \begin{tabular}{@{}lccccc@{}}
    \toprule
              & GC     & LC     & BLEU  & COMET  & Acc.  \\ \midrule
    \docmodel & $\varnothing$      & \cmark & 25.01 & 0.3021 & 66.99 \\ 
             & \xmark & \xmark & ---     & ---      & ---     \\ \midrule
    \ourscon  & \cmark & \cmark & \textbf{25.31} & \textbf{0.3173} & \textbf{70.92} \\
              & \xmark & \cmark & 24.86 & 0.2738 & 66.83 \\ 
              & \xmark & \xmark & ---     & ---      & ---     \\ \midrule
    \oursdis  & \cmark & \cmark & \textbf{25.41} & \textbf{0.3101} & \textbf{72.04} \\
              & \xmark & \cmark & 24.81 & 0.2037 & 69.30 \\ 
              & \xmark & \xmark & ---     & ---      & ---     \\ \bottomrule
    \end{tabular}

    \caption{
        Ablation study for data shuffling on \dataset{TED}.
        \cmark indicates the golden context.
        \xmark indicates the irrelevant context.
        --- indicates the model fails to converge.
        $\varnothing$ indicates \docmodel is not associated with GC.
        The best results for \ourscon and \oursdis are highlighted in \textbf{bold} respectively.
    }
    \label{tab:shuffle_ted}
\end{table}

\paragraph{Ablation Study for Data Shuffling}
To preserve the linguistic structure of the original document, we do not shuffle examples during training.
If the examples are shuffled, the predictions of the current sentence are conditioned on the gold local context (LC) and the irrelevant global context (GC).
In this section, we investigate how the data shuffling affects \ours.
We present the results in \autoref{tab:shuffle_ted}.
We observe the performance reduction for \ourscon and \oursdis when the current sentence is conditioned on the gold LC but irrelevant GC.
When conditioned on the irrelevant GC, \ours even performs worse than \docmodel which is free from the irrelevant GC.
We also train \ours and \docmodel with the completely irrelevant context and find out that both models fail to converge.
Hence, we confirm that the relatedness between the context and current sentence is of vital importance in DocNMT and \ours can effectively leverage the information from the context beyond the scope of the pseudo-documents.

%% file: 6_analysis.tex
\section{Analysis}

In this section, we investigate the effectiveness of \ours on the contextual awareness and the quality saturation problem.
We also demonstrate how the inference batch size affects the model predictions.
A visualization of FBA attention map is presented in \autoref{sec:viz_fba}.

\begin{table}[t]
    \small
    \centering
    \begin{tabular}{@{}lcccc@{}}
        \toprule
              & avg            & \textit{er}    & \textit{es}    & \textit{sie}   \\ \midrule
    \docmodel & 66.99          & 56.82          & 89.20          & 54.95          \\
    \mbe      & 68.12          & 52.57          & 89.72          & 62.07          \\
    \abdmodel & 68.25          & 55.30          & \textbf{90.65} & 58.82          \\ \midrule
    \ourscon  & 70.92          & 56.65          & 89.52          & \textbf{66.60} \\
    \oursdis  & \textbf{72.04} & \textbf{60.02} & 89.67          & 66.42          \\ \bottomrule
    \end{tabular}
    \caption{
        Accuracy (in \%) on the contrastive test set for \dataset{TED} with regard to the anaphoric pronoun types.
        The best results are highlighted in \textbf{bold}.
    }
    \label{tab:disco_ted}
\end{table}

\paragraph{Contextual Awareness}
In English-German translation, the choice of anaphoric pronoun types, including feminine \textit{sie}, neutral \textit{er} and masculine \textit{es}, commonly depends on its context.
We present the accuracy with regard to the anaphoric pronoun types given by the selected models trained on \dataset{TED} in \autoref{tab:disco_ted}.
\oursdis is the only approach that demonstrates substantial improvements on the neutral \textit{er}.
For the feminine \textit{sie}, \ourscon and \oursdis both outperform \docmodel by approximately 12\% accuracy.
\mbe and \abdmodel only improves the accuracy on the feminine \textit{sie} by 7\% and 4\% respectively.
We also present the change of accuracy given by the selected models against \docmodel with regard to the antecedent distance on \dataset{TED} in \autoref{fig:distance_ted}.
Compared with \docmodel, the approaches that leverage the batch-level information all effectively improves the accuracy on the distant context (antecedent distance $\geq 2$).
\ourscon significantly outperforms \docmodel with regard to the accuracy on the distant context by more than 8\%, while \oursdis outperforms \docmodel by more than 10\% on the distant context.
All these results demonstrate that \ours can effectively improve the contextual awareness on the discourse phenomena.

\begin{figure}[t]
    \centering
    \begin{tikzpicture}[scale=0.5]
        \begin{axis}[
            width  = 0.85*\textwidth,
            height = 6.5cm,
            major x tick style = transparent,
            ybar=2,
            bar width=6pt,
            ymajorgrids = true,
            grid style={line width=.1pt, draw=gray!50},
            ylabel = {\LARGE $\Delta_{\textrm{Acc.}}$},
            xlabel = {\LARGE Antecedent Distance},
            symbolic x coords={$0$, $1$, $2$, $3$, $>3$},
            xtick = data,
            scaled y ticks = false,
            enlarge x limits=0.2,
            ymin=-2,
            legend cell align=left,
            legend columns=4,
            legend style={
                    at={(1,1)},
                    anchor=south east,
                    column sep=1ex
            }
        ]
    
            \addplot[style={gray,fill=gray,mark=none}]
                    coordinates { ($0$,1.25) ($1$,0.23) ($2$,3.78) ($3$,3.66) ($>3$,2.72)};
    
            \addplot[style={orange,fill=orange,mark=none}]
                    coordinates { ($0$,-1.08) ($1$,0.82) ($2$,4.90) ($3$,5.58) ($>3$,3.17)};
    
            \addplot[style={blue,fill=blue,mark=none}]
                    coordinates { ($0$,1.79) ($1$,3.22) ($2$,8.68) ($3$,8.90) ($>3$,4.53)};

            \addplot[style={red,fill=red,mark=none}]
                    coordinates { ($0$,2.46) ($1$,4.14) ($2$,11.59) ($3$,10.30) ($>3$,4.75)};

            \legend{\mbe, \abdmodel, \ourscon, \oursdis}
        \end{axis}
    \end{tikzpicture}
    
    \caption{
        The change of accuracy (in \%; $\Delta_{\textrm{acc}}$) given by the selected models against \docmodel with regard to the antecedent distance (in sentences) on \dataset{TED}.
    }
    \label{fig:distance_ted}
\end{figure}
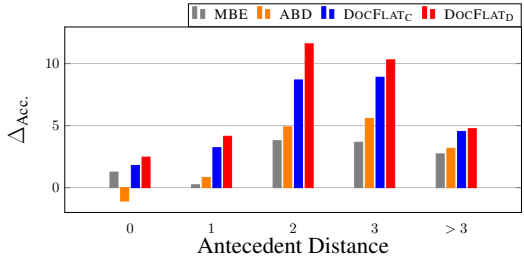

\begin{figure}[t]  
    \centering 
    \begin{subfigure}[c]{0.45\linewidth}
        \centering
        \begin{tikzpicture}[scale=0.4]
            \begin{axis}[
                xlabel={\LARGE Local Context Size},
                ylabel={\LARGE BLEU},
                y label style={at={(-0.05,0.5)}},
                xtick={1,2,3,4,5},
                xticklabels={1,2,3,5,10},
                ymajorgrids = true,
                grid style={line width=.1pt, draw=gray!50},
                y tick label style={
                    /pgf/number format/.cd,
                    fixed,
                    fixed zerofill,
                    precision=2
                },
                ymin=24.7, ymax=26.2,
                legend style={
                    at={(1,1)},
                    anchor=south east,
                    column sep=0ex,
                    font=\scriptsize,
                    legend columns=-1,
                    legend cell align=left,
                }
            ]
            \addplot[color=orange,mark=square] coordinates {
                (1,25.26)
                (2,25.17)
                (3,25.01)
                (4,25.06)
                (5,24.96)
            };
            \addplot[color=blue,mark=*] coordinates {
                (1,25.58)
                (2,25.49)
                (3,25.51)
                (4,25.77)
                (5,25.81)
            };
            \addplot[color=red,mark=o] coordinates {
                (1,25.17)
                (2,25.62)
                (3,25.61)
                (4,25.68)
                (5,25.52)
            };
            \legend{\docmodel, \ourscon, \oursdis}
            \end{axis}  
        \end{tikzpicture}
    \end{subfigure}
    \hspace{1mm}
    \begin{subfigure}[c]{0.45\linewidth}
        \centering
        \begin{tikzpicture}[scale=0.4] 
            \begin{axis}[
                xlabel={\LARGE Local Context Size},
                ylabel={\LARGE COMET},
                y label style={at={(-0.05,0.5)}},
                xtick={1,2,3,4,5},
                xticklabels={1,2,3,5,10},
                ymin=0.285, ymax=0.335,
                ymajorgrids = true,
                grid style={line width=.1pt, draw=gray!50},
                y tick label style={
                    /pgf/number format/.cd,
                    fixed,
                    fixed zerofill,
                    precision=2
                },
                legend style={
                    at={(1,1)},
                    anchor=south east,
                    column sep=0ex,
                    font=\scriptsize,
                    legend columns=-1,
                    legend cell align=left,
                }
            ]
            \addplot[color=orange,mark=square] coordinates {
                (1,0.3053)
                (2,0.3074)
                (3,0.3021)
                (4,0.2968)
                (5,0.2942)
            };
            \addplot[color=blue,mark=*] coordinates {
                (1,0.3212)
                (2,0.3210)
                (3,0.3173)
                (4,0.3216)
                (5,0.3187)
            };
            \addplot[color=red,mark=o] coordinates {
                (1,0.3089)
                (2,0.3163)
                (3,0.3101)
                (4,0.3189)
                (5,0.3127)
            };
            \legend{\docmodel, \ourscon, \oursdis}
            \end{axis} 
        \end{tikzpicture}
    \end{subfigure}
    \caption{
        BLEU (left) and COMET (right) against the LC size of pseudo-document (in sentences) by \docmodel, \ourscon and \oursdis on \dataset{TED}.
    }
    \label{fig:ctx_size}
\end{figure}
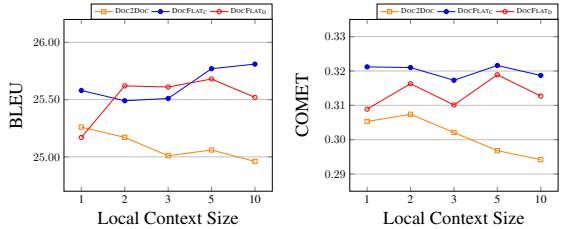

\paragraph{Effect of \ours on Quality Saturation}
\docmodel suffers from the \textit{quality saturation problem} as shown in \autoref{fig:doc_ctx_size}.
We investigate if \ours also suffers from the same problem.
We display the results in \autoref{fig:ctx_size} and observe that \ourscon and \oursdis perform consistently with regard to the LC size.
We conjecture the reasons for this observation from two perspectives.
When the LC size is small, the information from GC introduced by FBA complements the missing information in LC.
When the LC size is large enough, most information from GC is already covered by LC and FBA functions as a regularizer.

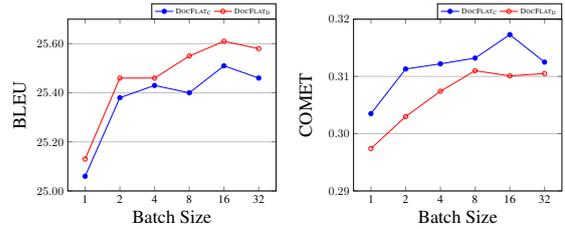
\begin{figure}[t]  
    \centering 
    \begin{subfigure}[c]{0.45\linewidth}
        \centering
        \begin{tikzpicture}[scale=0.4]
            \begin{axis}[
                xlabel={\LARGE Batch Size},
                ylabel={\LARGE BLEU},
                y label style={at={(-0.05,0.5)}},
                xtick={1,2,3,4,5,6},
                xticklabels={1,2,4,8,16,32},
                xmin=0.5, xmax=6.5,
                ymin=25, ymax=25.7,
                ymajorgrids = true,
                grid style={line width=.1pt, draw=gray!50},
                y tick label style={
                    /pgf/number format/.cd,
                    fixed,
                    fixed zerofill,
                    precision=2
                },
                legend style={
                    at={(1,1)},
                    anchor=south east,
                    column sep=0ex,
                    font=\scriptsize,
                    legend columns=-1,
                    legend cell align=left,
                }
            ]
            \addplot[color=blue,mark=*] coordinates {
                (1,25.06)
                (2,25.38)
                (3,25.43)
                (4,25.40)
                (5,25.51)
                (6,25.46)
            };

            \addplot[color=red,mark=o] coordinates {
                (1,25.13)
                (2,25.46)
                (3,25.46)
                (4,25.55)
                (5,25.61)
                (6,25.58)
            };
            \legend{\ourscon, \oursdis};

            \end{axis}  
        \end{tikzpicture}
    \end{subfigure}
    \hspace{1mm}
    \begin{subfigure}[c]{0.45\linewidth}
        \centering
        \begin{tikzpicture}[scale=0.4] 
            \begin{axis}[
                xlabel={\LARGE Batch Size},
                ylabel={\LARGE COMET},
                y label style={at={(-0.05,0.5)}},
                xtick={1,2,3,4,5,6},
                xticklabels={1,2,4,8,16,32},
                xmin=0.5, xmax=6.5,
                ymin=0.29, ymax=0.32,
                ymajorgrids = true,
                grid style={line width=.1pt, draw=gray!50},
                y tick label style={
                    /pgf/number format/.cd,
                    fixed,
                    fixed zerofill,
                    precision=2
                },
                legend style={
                    at={(1,1)},
                    anchor=south east,
                    column sep=0ex,
                    font=\scriptsize,
                    legend columns=-1,
                    legend cell align=left,
                }
            ]
            \addplot[color=blue,mark=*] coordinates {
                (1,0.3035)
                (2,0.3113)
                (3,0.3122)
                (4,0.3132)
                (5,0.3173)
                (6,0.3125)
            };
            \addplot[color=red,mark=o] coordinates {
                (1,0.2974)
                (2,0.3030)
                (3,0.3074)
                (4,0.3110)
                (5,0.3101)
                (6,0.3105)
            };
            \legend{\ourscon, \oursdis};

            \end{axis} 
        \end{tikzpicture}
    \end{subfigure}
    \caption{
        BLEU (left) and COMET (right) against the inference batch size (in sentences) given by \ourscon and \oursdis on \dataset{TED}.
    }
    \label{fig:infer_batch_size}
\end{figure}

\paragraph{Inference Batch Size}
At the inference stage, \ours is also able to leverage the batch-level information.
We visualize how the inference batch size impacts the model performance in \autoref{fig:infer_batch_size}.
Overall, the model performance of \ours is positively correlated to the inference batch size.
When the batch size is 1, \ourscon and \oursdis still outperform \docmodel, suggesting the FBA can help the model utilize the distant context during training.
The performance gain on BLEU and COMET for both \ourscon and \oursdis diminishes as the inference batch size increases, and we do not observe further improvement when the inference batch size is larger than 16, suggesting the over-distant context is less influential to the predictions of the current sentence.

%% file: 7_related_work.tex
\section{Related Work}

\paragraph{Document-Level NMT}
Numerous document-level NMT approaches have been proposed in recent years.
\citet{tiedemann-scherrer-2017-neural} firstly proposed the simple concatenation-based DocNMT model.
Existing works in the document-level NMT widely spread on a variety of research topics, including the model architecture \cite{miculicich-etal-2018-document,maruf-etal-2019-selective,zhang-etal-2021-multi}, training methods \cite{sun-etal-2022-rethinking, lei-etal-2022-codonmt}, evaluation \cite{bawden-etal-2018-evaluating,jiang-etal-2022-blonde}, etc.
\citet{zhang-etal-2018-improving} incorporate the contextual information using an independent context encoder.
\citet{bao-etal-2021-g} propose group attention that introduce a locality bias to force the model to focus on the recent context.
\citet{morishita-etal-2021-context} compute the average representation of all the source tokens, which is the only close work to ours.
\citet{DBLP:journals/csur/MarufSH21} present a detailed review on DocNMT.

\paragraph{Batch-Level Information}
Modeling instance relationships in the batch is relatively less explored.
Prior works leveraging the instance relationships are mostly from the computer vision area.
\citet{DBLP:conf/icml/IoffeS15} keep the running mean and variance in the batch to normalize the training and testing instances.
\citet{DBLP:conf/iclr/ZhangCDL18} linearly combine a random pair of instances to improve the model generalization.
\citet{DBLP:conf/bmvc/MondalJS21} use graph neural networks to aggregate information from similar images.
\citet{Hou_2022_CVPR} propose \model{BatchFormer} to improve the long-tail recognition by combining different instances.
Our work is directly inspired by \citet{DBLP:conf/nips/KossenBLGRG21} that computes the pairwise similarity among all the batched instances, with distinct motivation.
We aim to utilize distant context beyond the pseudo-document boundaries, instead of finding the similar patterns.

%% file: 8_conclusion.tex
\section{Conclusion}

In this work, we address the limitation of the pseudo-document formation in the DocNMT by utilizing the batch-level information.
We propose a novel Document Flattening (\ours) technique that integrates \myattn (FBA) and \gate (NCG) into the Transformer model.
FBA enables the current sentence to access the information beyond the pseudo-document boundaries and NCG identifies the usefulness of context and controls the information flow. 
We conduct comprehensive experiments and analyses on three benchmark datasets for English-German translation.
We demonstrate that \ours outperforms several strong baselines with statistical significance.
The analyses highlight that \ours can effectively alleviate the quality saturation problem in DocNMT and capture the long-range information.

%% file: 9_limitation.tex
\section{Limitation}

As suggested in \autoref{fig:infer_batch_size}, the performance of \ours is positively correlated to the inference batch size. 
This is because large inference batch size could help \ours to better utilize distant context within the same inference batch. 
However, this property of \ours could become an issue when there are only limited inference computational resources available.

\section*{Acknowledgment}
This research is supported by the ARC Future Fellowship FT190100039. 
This work is partly sponsored by the Air Force Research Laboratory and DARPA under agreement number FA8750-19-2-0501. 
The U.S. Government is authorized to reproduce and distribute reprints for Governmental purposes notwithstanding any copyright notation thereon. 
The authors are grateful to the anonymous reviewers for their helpful comments to improve the manuscript.

%% file: appendix.tex
\section{A Concrete Example at the Source Side}
\label{sec:src_example}

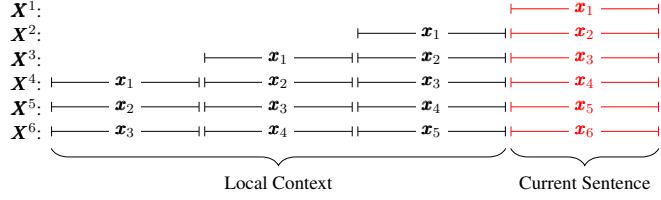
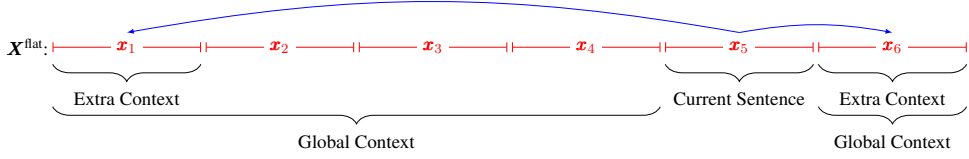
\begin{figure*}[t]
    
    \begin{subfigure}{\textwidth}
        \centering
        \scalebox{0.65}{
            \begin{tikzpicture}

                \node[] at (-0.5, 0) () {$\vX^{1}$:} ;
                \draw[|-|, red] (9.3,0) --  node[fill=white] {$\vx_{1}$} (12.3,0);
    
                \node[] at (-0.5, -0.5) () {$\vX^{2}$:} ;
                \draw[|-|] (6.2,-0.5) --  node[fill=white] {$\vx_{1}$} (9.2,-0.5);
                \draw[|-|, red] (9.3,-0.5) --  node[fill=white] {$\vx_{2}$} (12.3,-0.5);
    
                \node[] at (-0.5, -1) () {$\vX^{3}$:} ;
                \draw[|-|] (3.1,-1) --  node[fill=white] {$\vx_{1}$} (6.1,-1);
                \draw[|-|] (6.2,-1) --  node[fill=white] {$\vx_{2}$} (9.2,-1);
                \draw[|-|, red] (9.3,-1) --  node[fill=white] {$\vx_{3}$} (12.3,-1);
    
                \node[] at (-0.5, -1.5) () {$\vX^{4}$:} ;
                \draw[|-|] (0,-1.5) --  node[fill=white] {$\vx_{1}$} (3,-1.5);
                \draw[|-|] (3.1,-1.5) --  node[fill=white] {$\vx_{2}$} (6.1,-1.5);
                \draw[|-|] (6.2,-1.5) --  node[fill=white] {$\vx_{3}$} (9.2,-1.5);
                \draw[|-|, red] (9.3,-1.5) --  node[fill=white] {$\vx_{4}$} (12.3,-1.5);
    
                \node[] at (-0.5, -2) () {$\vX^{5}$:} ;
                \draw[|-|] (0,-2) --  node[fill=white] {$\vx_{2}$} (3,-2);
                \draw[|-|] (3.1,-2) --  node[fill=white] {$\vx_{3}$} (6.1,-2);
                \draw[|-|] (6.2,-2) --  node[fill=white] {$\vx_{4}$} (9.2,-2);
                \draw[|-|, red] (9.3,-2) --  node[fill=white] {$\vx_{5}$} (12.3,-2);
    
                \node[] at (-0.5, -2.5) () {$\vX^{6}$:} ;
                \draw[|-|] (0,-2.5) --  node[fill=white] {$\vx_{3}$} (3,-2.5);
                \draw[|-|] (3.1,-2.5) --  node[fill=white] {$\vx_{4}$} (6.1,-2.5);
                \draw[|-|] (6.2,-2.5) --  node[fill=white] {$\vx_{5}$} (9.2,-2.5);
                \draw[|-|, red] (9.3,-2.5) --  node[fill=white] {$\vx_{6}$} (12.3,-2.5);

                \draw [decorate,decoration={brace,amplitude=10pt,mirror,raise=2ex}]
                    (0,-2.5) -- (9.2,-2.5) node[midway,yshift=-30pt]{Local Context};
                \draw [decorate,decoration={brace,amplitude=10pt,mirror,raise=2ex}]
                    (9.3,-2.5) -- (12.3,-2.5) node[midway, yshift=-30pt]{Current Sentence};

            \end{tikzpicture}
        }
        \caption{
            An example batch of pseudo-documents $\vB_{\textrm{src}}=\{ \vX^{1}, \vX^{2}, \vX^{3}, \vX^{4}, \vX^{5}, \vX^{6} \}$ at the source side.
            Each $\vX^{j}$ contains four consecutive sentences and $\vx_{i}$ indicates the $i$-th sentence of the same original document.
            \textcolor{red}{The segments in red} indicate the current sentence of each pseudo-document.
        }
        \label{fig:src_pseudo_docs}
    \end{subfigure}
    \vfill
    \begin{subfigure}{\textwidth}
        \centering
        \scalebox{0.65}{
            \begin{tikzpicture}
                \node[] at (-0.5, 0) () {$\vX^{\textrm{flat}}$:} ;
                \draw[|-|,red] (0,0) --  node[fill=white] {$\vx_{1}$} (3,0);
                \draw[|-|,red] (3.1,0) --  node[fill=white] {$\vx_{2}$} (6.1,0);
                \draw[|-|,red] (6.2,0) --  node[fill=white] {$\vx_{3}$} (9.2,0);
                \draw[|-|,red] (9.3,0) --  node[fill=white] {$\vx_{4}$} (12.3,0);
                \draw[|-|,red] (12.4,0) --  node[fill=white] {$\vx_{5}$} (15.4,0);
                \draw[|-|,red] (15.5,0) --  node[fill=white] {$\vx_{6}$} (18.5,0);
    
                \draw [-latex,blue] (13.9,0.3) to [out=170,in=10] (1.5,0.3);
                \draw [-latex,blue] (13.9,0.3) to [out=10,in=170] (17,0.3);

                \draw [decorate,decoration={brace,amplitude=10pt,mirror,raise=2ex}]
                    (0,0) -- (3,0) node[midway,yshift=-30pt]{Extra Context};
                \draw [decorate,decoration={brace,amplitude=10pt,mirror,raise=7ex}]
                    (0,0) -- (12.3,0) node[midway,yshift=-55pt]{Global Context};
                \draw [decorate,decoration={brace,amplitude=10pt,mirror,raise=2ex}]
                    (12.4,0) -- (15.4,0) node[midway,yshift=-30pt]{Current Sentence};
                \draw [decorate,decoration={brace,amplitude=10pt,mirror,raise=2ex}]
                    (15.5,0) -- (18.5,0) node[midway,yshift=-30pt]{Extra Context};
                \draw [decorate,decoration={brace,amplitude=10pt,mirror,raise=7ex}]
                    (15.5,0) -- (18.5,0) node[midway,yshift=-55pt]{Global Context};
    
            \end{tikzpicture}
        }
        \caption{
            An example of the flattened sequence $\vX^{\textrm{flat}}$ transformed from $\vB_{\textrm{tgt}}$ with \myattn.
            For the current sentence $\vx_{5}$, \textcolor{blue}{The blue arrows} indicate the extra inter-sentential attention for $\vx_{5}$ that our approach can model.
            $\vx_{1}$ and $\vx_{6}$ are the extra context introduced by our approach. 
        }
        \label{fig:src_flattenseq}
    \end{subfigure}
    \caption{
        An example batch of pseudo-documents at the source side and its flattened sequence.
    }
    \label{fig:src_exmaple}
\end{figure*}

We present a concrete example at the source side in \autoref{fig:src_exmaple}.

\section{Optimization and Hyperparameters}
\label{sec:hyperparam}

We use a two-stage training routine following the previous works \cite{zhang-etal-2018-improving, voita-etal-2019-good, lopes-etal-2020-document, bao-etal-2021-g}.
\paragraph{Stage I} 
We first train a \sentmodel NMT model.
The model is randomly initialized and optimized with Adam \cite{DBLP:journals/corr/KingmaB14} with $\beta_1=0.9$, $\beta_2=0.98$ and the learning rate $\alpha=5 \times 10^{-4}$.
The model is trained with the batch size of $32K$ tokens for both datasets and the dropout rate $p=0.3$.
The batch size of $32K$ tokens is achieved by using the batch size of $4096$ tokens and updating the model for every $8$ batches.
The learning rate schedule is the same as described in \citet{DBLP:conf/nips/VaswaniSPUJGKP17} with $4K$ warmup steps.
We use early stopping on validation loss.

\paragraph{Stage II}
The document-level models are all fine-tuned from the best \sentmodel model in the Stage I.
With the same learning rate schedule as the Stage I, we set the learning rate $\alpha=2 \times 10^{-4}$.
All the other hyperparameters are identical.
Training is early stopped on validation loss, and we average the last 5 checkpoints to report the model performance, following \citet{DBLP:conf/nips/VaswaniSPUJGKP17}.
Following \citet{bao-etal-2021-g}, we apply word dropout \cite{DBLP:conf/nips/GalG16, sennrich-etal-2016-edinburgh} to the inputs with $p=0.1$.

\section{\dataattn}
\label{sec:abd}

We adapt \dataattn (ABD) proposed by \citet{DBLP:conf/nips/KossenBLGRG21} to the DocNMT.
The model architecture is identical to \ours as shown in \autoref{fig:model} with FBA replaced with ABD.
Given a batch of hidden representations $\vH \in \mathbb{R}^{n \times d \times e}$, ABD is defined as follows:
\begin{align}
    \label{eq:abd}
    \begin{split}
        \tilde{\vH}_{\textrm{avg}} &= \textrm{AvgPool}(\vH) \in \mathbb{R}^{n \times 1 \times e}, \\
        \tilde{\vH}_{\textrm{flat}} &= \textrm{Flatten}(\tilde{\vH}_{\textrm{avg}}) \in \mathbb{R}^{1 \times n \times e}, \\
        \tilde{\vH}_{\textrm{mhsa}} &= \textrm{MHSA}(\tilde{\vH}_{\textrm{flat}}, \tilde{\vH}_{\textrm{flat}}, \tilde{\vH}_{\textrm{flat}}) \in \mathbb{R}^{1 \times n \times e}, \\
        \tilde{\vH}_{\textrm{rsh}} &= \textrm{Reshape}(\textrm{Repeat}(\tilde{\vH}_{\textrm{mhsa}})) \in \mathbb{R}^{n \times d \times e}, \\
        \tilde{\vH} &= \textrm{LN}(\vH + \tilde{\vH}_{\textrm{rsh}}) \in \mathbb{R}^{n \times d \times e}.
    \end{split}
\end{align}
There is a noticeable difference in \autoref{eq:abd} from \autoref{eq:myattn} that we apply the average pooling to the sequence to obtain the instance representation, instead of directly flattening the token representations into a single vector.
ABD is originally designed for fixed-length data, and it is non-trivial to apply ABD to the variable-length inputs, and hence, we use the average pooling for simplicity.

We present the preliminary study of \abdmodel on \dataset{TED} in \autoref{tab:abd}.
When ABD is applied at the decoder side, the model performance is significantly reduced.
This observation suggests that the linguistic structure at the target side is of vital importance to DocNMT.

\begin{table}[t]
    \small
    \centering
    \begin{tabular}{@{}lccccc@{}}
    \toprule
              & Enc.   & Dec.   & BLEU  & COMET   & Acc.  \\ \midrule
    \docmodel & \xmark & \xmark & 25.01 & 0.3021  & 66.99 \\ \midrule
    \abdmodel &        &        & 18.57 & \llap{-}0.1202 & 66.55 \\
              & \xmark &        & 18.46 & \llap{-}0.1123 & 66.47 \\
              &        & \xmark & 24.97 & 0.3046  & 68.25 \\ \bottomrule
    \end{tabular}
    \caption{
        Preliminary study on the usage of ABD on \dataset{TED}.
        \xmark indicates \abdmodel is removed.
    }
    \label{tab:abd}
\end{table}

\section{Visualization of FBA}
\label{sec:viz_fba}

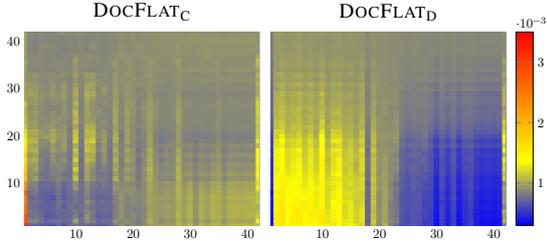
\begin{figure}[t]
    \centering 
    \begin{tikzpicture}[scale=0.45]
        \begin{groupplot}[
            group style={
                group size=2 by 1,
                xlabels at=edge top,
                x descriptions at=edge bottom,
                y descriptions at=edge left,
                horizontal sep=10pt,
            },
            xmin=1, xmax=42,
            ymin=1, ymax=42,
            point meta min=0.0003,
            point meta max=0.0035,
        ]
        \nextgroupplot[
            title={\LARGE \ourscon},
        ]
        \addplot [matrix plot*,point meta=explicit] file [] {attn_avg_con_new.dat};

        \nextgroupplot[
            title={\LARGE \oursdis},
            colorbar,
        ]
        \addplot [matrix plot*,point meta=explicit] file [] {attn_avg_dis_new.dat};

        \end{groupplot}
    
    \end{tikzpicture}
    \caption{
        Sentence-wise attention map produced by the FBA of \ourscon and \oursdis at the encoder side.
        $x$-axis indicates sentences as the keys of FBA.
        $y$-axis indicates sentences as the queries of FBA.
    }
    \label{fig:attn_map}
\end{figure}

To better understand the behavior of FBA, we visualize the sentence-wise attention map learned by the FBA of \ourscon and \oursdis at the encoder side in \autoref{fig:attn_map}.
The sample document for producing \autoref{fig:attn_map} can be found in \autoref{tab:sample_doc}.

It is infeasible to visualize the token-wise attention map for a very long sequence, so we aggregate the token-wise attention scores into the sentence-level.
We denote the token-wise attention map for the flattened sequence as $\mathcal{A}$.
For each pair of sentences $\vs_i$ attending to $\vs_j$, their token-wise attention map is a patch of $\mathcal{A}$, denoted as $\mathcal{A}^{p}_{ij}$.
We aggregate the token-level attention scores in the attention patch $\mathcal{A}^{p}_{ij}$ into a sentence-level score, as follows:
\begin{align}
    \label{eq:sent_attn}
    \mathcal{A}_{\mathcal{S}}(i, j) = \frac{1}{\left\lvert \mathcal{A}^{p}_{ij} \right\rvert } \sum \mathcal{A}^{p}_{ij}
\end{align}
where $\left\lvert \mathcal{A}^{p}_{ij} \right\rvert$ is the size of $\mathcal{A}^{p}_{ij}$ and $\mathcal{A}_{\mathcal{S}}(i, j)$ is the sentence-level attention score for $\vs_i$ attending to $\vs_j$.

The FBA of \ourscon at the encoder side considers all the sentences in the document to be equally important, while the one of \oursdis approximately splits the whole documents into two parts.
As shown by \oursdis in \autoref{fig:attn_map}, sentences in the first half focus more on its neighbors in the same split but those in the second half roughly attend to all the sentences in the documents.
This observation implies that the latter context is more dependent on the former context.

\begin{table*}[t]
    \tiny
    \centering
    \begin{tabular}{@{}lp{0.8\textwidth}@{}}
    \toprule
    idx & Context                                                                                                                                                                                                                                                                                 \\ \midrule
    1   & <d>                                                                                                                                                                                                                                                                                      \\
    2   & We're at a tipping point in human history, a species poised between gaining the stars and losing the planet we call home.                                                                                                                                                                \\
    3   & Even in just the past few years, we've greatly expanded our knowledge of how Earth fits within the context of our universe.                                                                                                                                                              \\
    4   & NASA's Kepler mission has discovered thousands of potential planets around other stars, indicating that Earth is but one of billions of planets in our galaxy.                                                                                                                           \\
    5   & Kepler is a space telescope that measures the subtle dimming of stars as planets pass in front of them, blocking just a little bit of that light from reaching us.                                                                                                                       \\
    6   & Kepler's data reveals planets' sizes as well as their distance from their parent star.                                                                                                                                                                                                   \\
    7   & Together, this helps us understand whether these planets are small and rocky, like the terrestrial planets in our own Solar System, and also how much light they receive from their parent sun.                                                                                          \\
    8   & In turn, this provides clues as to whether these planets that we discover might be habitable or not.                                                                                                                                                                                     \\
    9   & Unfortunately, at the same time as we're discovering this treasure trove of potentially habitable worlds, our own planet is sagging under the weight of humanity.                                                                                                                        \\
    10  & 2014 was the hottest year on record.                                                                                                                                                                                                                                                     \\
    11  & Glaciers and sea ice that have been with us for millennia are now disappearing in a matter of decades.                                                                                                                                                                                   \\
    12  & These planetary-scale environmental changes that we have set in motion are rapidly outpacing our ability to alter their course.                                                                                                                                                          \\
    13  & But I'm not a climate scientist, I'm an astronomer.                                                                                                                                                                                                                                      \\
    14  & I study planetary habitability as influenced by stars with the hopes of finding the places in the universe where we might discover life beyond our own planet.                                                                                                                           \\
    15  & You could say that I look for choice alien real estate.                                                                                                                                                                                                                                  \\
    16  & Now, as somebody who is deeply embedded in the search for life in the universe, I can tell you that the more you look for planets like Earth, the more you appreciate our own planet itself.                                                                                             \\
    17  & Each one of these new worlds invites a comparison between the newly discovered planet and the planets we know best: those of our own Solar System.                                                                                                                                       \\
    18  & Consider our neighbor, Mars.                                                                                                                                                                                                                                                             \\
    19  & Mars is small and rocky, and though it's a bit far from the Sun, it might be considered a potentially habitable world if found by a mission like Kepler.                                                                                                                                 \\
    20  & Indeed, it's possible that Mars was habitable in the past, and in part, this is why we study Mars so much.                                                                                                                                                                               \\
    21  & Our rovers, like Curiosity, crawl across its surface, scratching for clues as to the origins of life as we know it.                                                                                                                                                                      \\
    22  & Orbiters like the MAVEN mission sample the Martian atmosphere, trying to understand how Mars might have lost its past habitability.                                                                                                                                                      \\
    23  & Private spaceflight companies now offer not just a short trip to near space but the tantalizing possibility of living our lives on Mars.                                                                                                                                                 \\
    24  & But though these Martian vistas resemble the deserts of our own home world, places that are tied in our imagination to ideas about pioneering and frontiers, compared to Earth Mars is a pretty terrible place to live.                                                                  \\
    25  & Consider the extent to which we have not colonized the deserts of our own planet, places that are lush by comparison with Mars.                                                                                                                                                          \\
    26  & Even in the driest, highest places on Earth, the air is sweet and thick with oxygen exhaled from thousands of miles away by our rainforests.                                                                                                                                             \\
    27  & I worry -- I worry that this excitement about colonizing Mars and other planets carries with it a long, dark shadow: the implication and belief by some that Mars will be there to save us from the self-inflicted destruction of the only truly habitable planet we know of, the Earth. \\
    28  & As much as I love interplanetary exploration, I deeply disagree with this idea.                                                                                                                                                                                                          \\
    29  & There are many excellent reasons to go to Mars, but for anyone to tell you that Mars will be there to back up humanity is like the captain of the Titanic telling you that the real party is happening later on the lifeboats.                                                           \\
    30  & Thank you.                                                                                                                                                                                                                                                                               \\
    31  & But the goals of interplanetary exploration and planetary preservation are not opposed to one another.                                                                                                                                                                                   \\
    32  & No, they're in fact two sides of the same goal: to understand, preserve and improve life into the future.                                                                                                                                                                                \\
    33  & The extreme environments of our own world are alien vistas.                                                                                                                                                                                                                              \\
    34  & They're just closer to home.                                                                                                                                                                                                                                                             \\
    35  & If we can understand how to create and maintain habitable spaces out of hostile, inhospitable spaces here on Earth, perhaps we can meet the needs of both preserving our own environment and moving beyond it.                                                                           \\
    36  & I leave you with a final thought experiment: Fermi's paradox.                                                                                                                                                                                                                            \\
    37  & Many years ago, the physicist Enrico Fermi asked that, given the fact that our universe has been around for a very long time and we expect that there are many planets within it, we should have found evidence for alien life by now.                                                   \\
    38  & So where are they?                                                                                                                                                                                                                                                                       \\
    39  & Well, one possible solution to Fermi's paradox is that, as civilizations become technologically advanced enough to consider living amongst the stars, they lose sight of how important it is to safeguard the home worlds that fostered that advancement to begin with.                  \\
    40  & It is hubris to believe that interplanetary colonization alone will save us from ourselves, but planetary preservation and interplanetary exploration can work together.                                                                                                                 \\
    41  & If we truly believe in our ability to bend the hostile environments of Mars for human habitation, then we should be able to surmount the far easier task of preserving the habitability of the Earth.                                                                                    \\
    42  & Thank you.                                                                                                                                                                                                                                                                               \\ \bottomrule
    \end{tabular}
    \caption{
        The sample document used for producing \autoref{fig:attn_map}.
    }
    \label{tab:sample_doc}
\end{table*}